\RequirePackage{amsmath}
\documentclass[runningheads]{llncs}
\usepackage[T1]{fontenc}
\usepackage{graphicx}
\usepackage{xspace}
\usepackage{xcolor}
\usepackage{multirow}
\usepackage{enumitem}
\usepackage{booktabs}
\usepackage{amsmath,amssymb}

\definecolor{chromeyellow}{rgb}{1.0, 0.65, 0.0}

\newcommand{\realite}{\texttt{ReaLitE}\xspace}

\newcommand{\transerealite}{TransE+\realite\xspace}

\begin{document}
\title{ReaLitE: Enrichment of Relation Embeddings in Knowledge Graphs using Numeric Literals}

%
\titlerunning{ReaLitE: Relation Enrichment via Numeric Literals}
%
%
%
\author{Antonis Klironomos\inst{1,2} \and
Baifan Zhou\inst{3,4} \and
Zhuoxun Zheng\inst{1,3} \and
Gad-Elrab Mohamed\inst{1} \and
Heiko Paulheim\inst{2} \and
Evgeny Kharlamov\inst{1,3}}
\authorrunning{Klironomos et al.}
%
\institute{Bosch Center for Artificial Intelligence, Germany\\ \email{\{antonis.klironomos, mohamed.gad-elrab, evgeny.kharlamov\}@de.bosch.com}\\
\and
University of Mannheim, Germany\\
\email{heiko.paulheim@uni-mannheim.de}\\
\and
University of Oslo, Norway\\
\email{baifanz@ifi.uio.no}\\
\and
Oslo Metropolitan University, Norway
}
\maketitle              

\begin{abstract}

Most knowledge graph embedding (KGE) methods tailored for link prediction focus on the entities and relations in the graph, giving little attention to other literal values, which might encode important information. Therefore, some literal-aware KGE models attempt to either integrate numerical values into the embeddings of the entities or convert these numerics into entities during preprocessing, leading to information loss. Other methods concerned with creating relation-specific numerical features assume completeness of numerical data, which does not apply to real-world graphs. In this work, we propose \realite, a novel relation-centric KGE model that dynamically aggregates and merges entities' numerical attributes with the embeddings of the connecting relations. \realite is designed to complement existing conventional KGE methods while supporting multiple variations for numerical aggregations, including a learnable method.
We comprehensively evaluated the proposed relation-centric embedding using several benchmarks for link prediction and node classification tasks. The results showed the superiority of \realite\footnote{Pronounced as ``reality'', code: 
\url{https://github.com/boschresearch/ReaLitE}}
over the state of the art in both tasks. 



\end{abstract}

%

\section{Introduction}


\medskip
\noindent \textbf{Motivation.}
Knowledge graphs (KGs) represent information as interconnected entities and their relationships, typically structured in triples (\textit{head}, \textit{relation}, \textit{tail}). 
Moreover, KGs such as Wikidata~\cite{vrandecicWikidataFreeCollaborative2014} or YAGO~\cite{mahdisoltaniYAGO3KnowledgeBase2013} often incorporate various attributes with numerical values.
Knowledge graphs remain incomplete, lacking numerous links between entities. Consequently,  various \textit{knowledge graph embedding} (KGE) techniques, \textit{e.g.} TransE~\cite{NIPS2013_1cecc7a7} have been introduced to predict potential connections between entities. These embedding methods seek to represent the input KG by mapping entities and relations onto a lower-dimensional vector space. This process preserves existing triples in the KG while facilitating the prediction of new links. Subsequently, representations learned through KGEs have found application in diverse tasks, including node classification~\cite{portischKnowledgeGraphEmbedding2022}.



\begin{figure}[t]
    \centering
    \includegraphics[width=0.8\textwidth]{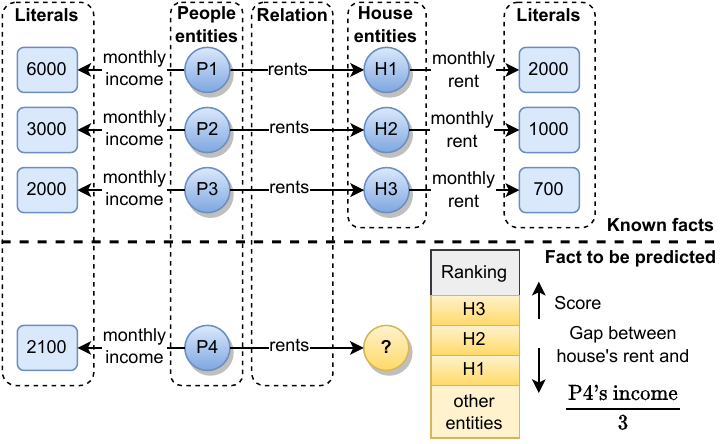}
    \caption{An example showing the potential benefit of the mathematical relationship between people's income and houses' rent during link prediction.}
    \label{fig:motivational-example}
    \vspace{-2ex}
\end{figure}

Each relation in a KG establishes a connection between pairs of entities.
Traditional KG embedding models, such as TransE~\cite{NIPS2013_1cecc7a7} and ComplEx~\cite{trouillonComplexEmbeddingsSimple2016},
neglect the literal attributes of entities. Recently, however, the field of multimodal KG completion has gained significant attention~\cite{shangLAFAMultimodalKnowledge2024,liuEntitiesLargeScaleMultiModal2024}. Also, the importance of relation representation is stressed by~\cite{longKGDMDiffusionModel2024a,liuEntitiesLargeScaleMultiModal2024}. At the same time, various types of relations may exhibit distinct patterns in how numerical attributes are linked to the head and tail of the relation.
For instance, a pattern may arise where \textit{people} (head) with a greater \textit{monthly income} (literal) tend to \textit{rent} (relation) \textit{houses} (tail) with higher \textit{monthly rent} (literal) (Figure~\ref{fig:motivational-example}).

By encoding the relationship between literals of head and tail entities, we can more accurately predict missing links for a head entity by intuitively ranking potential tail entities. The provided example shows that \textit{monthly income} $ \approx 3 \: *$ \textit{monthly rent} for the known triples. Thus, we would incorporate the monthly rent in the ranking of potential tail entities, using the difference to the head entity's
\textit{monthly income} / 3. 



\medskip
\noindent \textbf{State-of-the-art and Limitations.}
Existing KGE models that handle numerical literal values employ different strategies~\cite{geseseSurveyKnowledgeGraph2020a}.
The LiteralE model~\cite{kristiadiIncorporatingLiteralsKnowledge2019a}) incorporates literals into entity embeddings, while the KGA model~\cite{wangAugmentingKnowledgeGraphs2022c} converts them into entities during KG preprocessing. 
To our knowledge, KBLRN~\cite{garcia-duranKBLRNEndtoEndLearning2018d} is the only method attempting to use numeric literals as relation-specific features to enhance link prediction.
However, these methods overlook the potential patterns (exemplified in Figure~\ref{fig:motivational-example}) between different entity attributes for a given relation. KBLRN also faces challenges in incorporating absent literals, affecting the inclusion of numeric information. 
We argue that there exists an opportunity for improvement by integrating head and tail numeric information in a learnable manner, enabling the model to capture underlying numeric relationships. Since the relation acts as the connection point between head and tail entities, we propose using the relation embedding as a vessel for this combination.

In the realm of node classification, there is limited assessment regarding the performance of triple-scoring KGE methods (e.g., TransE). Additionally, numeric literals are frequently either excluded or handled as conventional nodes~\cite{ristoskiRDF2VecRDFGraph2016a,schlichtkrullModelingRelationalData2018,busbridgeRelationalGraphAttention2019}. A recent study proposed a literal preprocessing method based on the current state of the art and evaluated it in node classification~\cite{preisnerUniversalPreprocessingOperators2023a}. 




\medskip
\noindent \textbf{Approach.}
We introduce \realite, a relation-centric method to enhance vanilla KGE models by infusing numeric information from both head and tail entities into the embeddings of connecting relations. The values of each numeric attribute are aggregated separately for the relation's head and tail entities, generating two distinct numeric vectors. The aggregated numeric information is then combined with the relation embedding in a machine-learnable strategy. Such infusion enriches the relation embedding vectors with information about possible correlations between the numerical attributes of its head and tail entities. 

\medskip
\noindent \textbf{Contributions.}
Our contributions are summarized below: 
\begin{itemize}[topsep=3pt,parsep=0pt,partopsep=0pt,itemsep=2pt,leftmargin=*]
    \item We propose \realite, an approach that can be combined with any vanilla KGE method with relation embeddings. In addition, we demonstrate the integration of our method into existing KGE frameworks, highlighting its versatility and ease of adoption. 
    \item We experiment with different methods of aggregating numeric literals, including an automated method to learn a combination of multiple aggregation types. 
    \item We evaluate \realite extensively and compare it with state-of-the-art in two tasks: link prediction and node classification. For the former, we evaluate on the standard setting of link prediction, along with a more granular relation-focused evaluation. The results show that our approach is comparable or superior compared to the state-of-the-art methods, particularly on the numeric literals with higher correlation and on long-tail relations.

\end{itemize}

\section{Related Work}
\medskip
\noindent \textbf{Literal-aware Link Prediction.}
One of the early studies that focused on including numeric literals during link prediction introduced KBLRN: a model that consists of Latent, Relational, and Numerical Features~\cite{garcia-duranKBLRNEndtoEndLearning2018d}.
The inclusion of their model's numeric feature depends on the amount of missing numeric values in the dataset, which is a limitation.

LiteralE was one of the first methods to add literal information into vanilla KGE models by combining literals with the entity embeddings~\cite{kristiadiIncorporatingLiteralsKnowledge2019a}.
It does not provide a way to enhance the relation embeddings with numeric information directly.
This paper fills that gap and shows the benefits of fusing relation embeddings with numeric literals on link prediction and node classification tasks.

To the best of our knowledge, the latest method for handling numeric literals to perform link prediction is KGA (Knowledge Graph Augmentation)~\cite{wangAugmentingKnowledgeGraphs2022c}. KGA preprocesses the dataset to transform literals into entities.
While that study evaluates multiple models on link prediction, there is a lack of more granular evaluation and evaluation on downstream tasks. We first show the superiority of \realite in the typical evaluation setting for link prediction. Then, we investigate the models' performance across different types of relations. We observe significant performance gains favoring \realite for long-tail relations and relations with correlated head and tail attributes. 


\medskip
\noindent \textbf{Node Classification.}
Existing node classification models are mainly message passing (e.g. Graph Convolutional Networks (R-GCNs)~\cite{busbridgeRelationalGraphAttention2019,schlichtkrullModelingRelationalData2018,wilckeEndtoEndEntityClassification2020}) or feature extraction methods (e.g. RDF2Vec~\cite{ristoskiRDF2VecRDFGraph2016a}). Triple-scoring embedding methods (e.g., TransE) are rarely evaluated.

A recent study has highlighted 
that the link prediction performance of a KGE model may not necessarily translate to effectiveness in downstream tasks~\cite{zhangFineGrainedEvaluationKnowledge2020}. To assess KG embeddings, that study uses KG-based recommendation and question-answering downstream tasks. 
Our work focuses on node classification.

Another work that applies KGE models to downstream tasks 
uses product knowledge graphs (PKGs) in e-commerce, focusing on essential product relations for applications like marketing and recommendation~\cite{xuProductKnowledgeGraph2020a}.
Even though that paper evaluates triple-scoring KGE methods such as TransE on node classification, the dataset used (PKG) is inherently not suitable for these methods, as the authors explain in their paper. So, the performance of such methods in that dataset was expected and proven to be low.

Prior research with a more general scope evaluates traditional KGE methods on node classification using a particular KG~\cite{abboudNodeClassificationMeets2021}.
In our study, we use multiple datasets that also include literals.

As far as we know, in the downstream task of node classification, the latest study using triple-scoring KGE models with multimodal information in multi-relational KGs is~\cite{preisnerUniversalPreprocessingOperators2023a}.
The authors state that the study's handling of numerical data is partially based on \cite{wangAugmentingKnowledgeGraphs2022c} which is the current state of the art in literal-aware KGE methods. So, we choose to extend their study and show that \realite outperforms their method in node classification.
\section{Problem Description}

The formal representation of a KG with numeric literals can be expressed as a set of triples. Each triple consists of a head term ($h$), a relation or attribute ($r$ or $a$), and a tail term or literal value ($t$ or $v$). Officially, a KG can be defined as $G = \{(h, r, t) \mid h,t \in E, r \in R \} \cup \{(e, a, v) \mid e \in E, a \in A, v \in \mathbb{R}\}$. Here, $E$ represents the set of entities, $R$ is the set of relations, and $A$ is the set of data attributes.

Link prediction is a task that predicts the plausibility of a triple $(h, r, t)$ given a graph $G$.
In practice, this task is modeled as a ranking problem where for an input pair of $(h, r)$ the desired output is all entities $e_i \in E$ of $G$ ranked based on the probability $P$ of $(h, r, e_i)$ being a true triple. The ranks are computed using a scoring function $f(h, r, t)$ that produces scores proportional to $P$.

Another prominent downstream task is \textit{node classification}, which is concerned with predicting the category of a given entity. Formally, given a graph $G$, and a set of entity labels $Y$, this task aims to find a function $f(e_i) \rightarrow Y_i$, for each entity $e_i \in E$. A common way of adressing this task is to represent each entity by a pre-trained KGE $emb(e_i)$, and train a classifier $emb(e_i) \rightarrow Y_i$.

\section{Method: ReaLitE}
\label{sec:realite}

\begin{figure*}[t]
    \centering
    \includegraphics[width=1\textwidth]{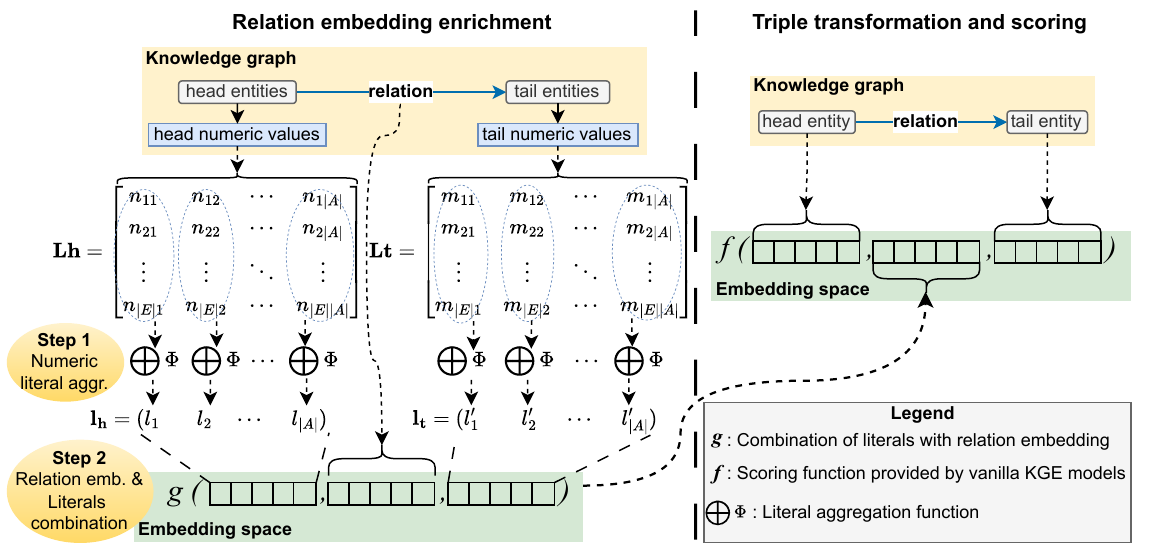}
    \caption{\realite's pipeline for relation embedding enrichment (left) given a triple to be scored (right).
    \realite aggregates the head and tail numeric literals (Step 1) and combines the aggregated literals with the relation embedding (Step 2).}
    \label{fig:model}
\end{figure*}

Our approach, \realite, involves a two-step procedure: (1) aggregating numeric literals on a per-relation basis and (2) integrating numeric literals into the relation embedding of vanilla KGE methods (Figure~\ref{fig:model}). 
These steps are independent of the scoring function, allowing our approach to complement existing KGE models as shown in the experiments.

Let $\mathbf{L} \in \mathbb{R}^{|E| \times |A|}$ be a matrix of literals, where $|E|$ is the number of entities and $|A|$ is the number of numeric attributes. Each entry $\mathbf{L}_{i,k}$ contains the literal value of the $k$-th numeric attribute for the $i$-th entity if a triple with the $i$-th entity and the $k$-th numeric attribute exists in the KG, and zero otherwise. $L$ is normalized using min-max normalization, as per the relevant literature~\cite{kristiadiIncorporatingLiteralsKnowledge2019a}.

For each relation, two matrices $\mathbf{Lh}, \mathbf{Lt} \in \mathbb{R
}^{|E|\times |A|}$ are created, containing the rows of $\mathbf{L}$ corresponding to the relation's head and tail entities, respectively. The remaining rows are zeroed out in $\mathbf{Lh}, \mathbf{Lt}$.
Then, an aggregation method is applied column-wise to $\mathbf{Lh}, \mathbf{Lt}$ to produce the vectors $\mathbf{l_h}, \mathbf{l_t} \in \mathbb{R}^{|A|}$. Depending on the configuration, this step uses either an aggregation type (\textit{e.g.}, $mean$) or a learnable linear combination of 11 different aggregation types.

The final part of \realite's architecture involves the function $g : \mathbb{R}^{|A|} \times \mathbb{R}^{D_r} \times \mathbb{R}^{|A|} \to \mathbb{R}^{D_r}$, where $D_r$ is the dimension of the relation embedding. The function $g$ takes as input: (a) the relation's embedding $r$ and (b) the two literal vectors $\mathbf{l_h}, \mathbf{l_t} \in \mathbb{R}^{|A|}$.
The output of $g$ is a vector of the same dimension as the relation embedding. The function $g$ is thoroughly described later in this section.

The resulting vector from $g$ is a literal-enhanced embedding vector, capable of substituting the original embedding vector of the relation within the scoring function.
So, every relation embedding $r_i$ is replaced with $\mathbf{r_i}^{lit} = g(\mathbf{l_h}, \mathbf{r_i}, \mathbf{l_t})$ in the scoring functions of respective model 
The remaining elements of the score functions remain unchanged.

Below we elaborate on the functionality of \realite by consecutively explaining both steps of the approach.

\subsection{Aggregation of Numeric Literals}
\label{sec:aggr-methods-desc}

To reduce the space complexity and number of trainable parameters of \realite,
we choose to convert the matrices $\mathbf{Lh}, \mathbf{Lt}$ to vectors. In the process, we consider it crucial to keep the number of columns intact, as it corresponds to the number of numeric attributes ($|A|$). This allows us to incorporate all the available numeric attributes and let the model decide which attributes are more useful than others during training.
So, we choose to aggregate the values of each numeric attribute (i.e., each column of $\mathbf{Lh}$ and $\mathbf{Lt}$) to create vectors $\mathbf{l_h}, \mathbf{l_t}$ (`Step 1' of Figure~\ref{fig:model}).
\begin{align}
    \mathbf{l_h} = & \begin{pmatrix} \bigoplus_{i=1}^{|E|} \Phi(\mathbf{Lh}_{i, 1}) & \cdots & \bigoplus_{i=1}^{|E|} \Phi(\mathbf{Lh}_{i, |A|}) \end{pmatrix} \label{eq:lh} \\
    \mathbf{l_t} = & \begin{pmatrix} \bigoplus_{i=1}^{|E|} \Phi(\mathbf{Lt}_{i, 1}) & \ \cdots & \bigoplus_{i=1}^{|E|} \Phi(\mathbf{Lt}_{i, |A|}) \end{pmatrix} \label{eq:lt}
\end{align}

We treat the aggregation method denoted by $\bigoplus \Phi$ as a tunable hyperparameter for \realite.
That is because depending on the skewness of the value distribution for each numeric attribute, the aggregation type that leads to better link prediction performance might vary.
The choices for $\bigoplus \Phi$ are commonly used measures of central tendency $\{mean, median, mode\}$ and other descriptive statistics $\{min, max, sum, count\}$, and measures of dispersion $\{variance, std, IQR, range\}$. Additionally, we provide the option to use a learnable linear combination $\mathbf{y}$ of all previous aggregation types (Eq.~\ref{aggr-combination}).
\begin{equation}
    \mathbf{y} =  \sigma(\mathbf{U} \cdot \mathbf{W}_a + \mathbf{b}_1) \label{aggr-combination} \\
\end{equation}
where, $\forall \phi \in $\{\textit{mean}, \textit{median}, \textit{mode}, \textit{min}, \textit{max}, \textit{sum}, \textit{count}, \textit{variance}, \textit{std}, \textit{IQR}, \textit{range}\}, $\exists$ row $\mathbf{u}$ of the matrix $\mathbf{U}^\intercal$ where:
\begin{equation}
    \mathbf{u} = \begin{pmatrix} \bigoplus_{i=1}^{|E|} \phi(\mathbf{L}_{i,1}) & \cdots & \bigoplus_{i=1}^{|E|} \phi(\mathbf{L}_{i,|A|}) \end{pmatrix} \label{aggr-apply}
\end{equation}

The input of linear function $\mathbf{y} \in \mathbb{R}^{|A| \times 1}$ is $\mathbf{U} \in \mathbb{R}^{|A| \times 11}$. 
The aggregation function is denoted by $\bigoplus \phi$. The learnable parameters are $\mathbf{W}_a \in \mathbb{R}^{11 \times 1}$ and $\mathbf{b} \in \mathbb{R}$, while $\sigma$ is the sigmoid function.
We apply the function in Eq.~\ref{aggr-combination} separately on $\mathbf{Lh}$ and $\mathbf{Lt}$ by substituting $\mathbf{L}$ for each of them in Eq.~\ref{aggr-apply}.

\subsection{Fusing Literals with Relation Embeddings}
\label{sec:function-g}
The aggregated head and tail numeric literals (vectors $\mathbf{l_h}$ [Eq.~\ref{eq:lh}] and $\mathbf{l_t}$ [Eq.~\ref{eq:lt}]) are combined with the relation embedding via the learnable function $g$ (`Step 2' of Figure~\ref{fig:model}). We provide two versions of this function: Linear (Eq.~\ref{eq:glin}) and Gated (Eq.~\ref{eq:ggated}). While the former uses a linear transformation, the latter employs a gated mechanism as done in~\cite{kristiadiIncorporatingLiteralsKnowledge2019a}, because the authors claim this mechanism lets $g$ use or ignore the numeric information as needed.
\begin{align}
    g_{lin} = & \mathbf{W}_r^T[\mathbf{l_h}, \mathbf{r}, \mathbf{l_t}] \label{eq:glin} + \mathbf{b}_2 \\
    g_{gated} = & \mathbf{z} \odot \mathbf{h} + (1-\mathbf{z}) \odot \mathbf{r} \label{eq:ggated}
\end{align}
\begin{align}
    \text{where: } & \mathbf{z} = \sigma(\mathbf{W}_{zlh}^T \mathbf{l_h} + \mathbf{W}_{zr}^T \mathbf{r} + \mathbf{W}_{zlt}^T \mathbf{l_t} + \mathbf{b}_3) \nonumber \\
     & \mathbf{h} =  h'(\mathbf{W}_r^T[\mathbf{l_h}, r, \mathbf{l_t}]) \nonumber
\end{align}

Except for the relation embedding $\mathbf{r} \in \mathbb{R}^{D_r}$, the trainable parameters are $\mathbf{W}_r \in \mathbb{R}^{|A| + D_r + |A| \times D_r}$, $\mathbf{W}_{zr} \in \mathbb{R}^{D_r \times D_r}$, $\mathbf{W}_{zlh},\mathbf{W}_{zlt} \in \mathbb{R}^{|A| \times D_r}$ and $\mathbf{b}_2,\mathbf{b}_3 \in \mathbb{R}^{D_r}$. 
With $\odot$ denoting element-wise multiplication, $\sigma$ as the sigmoid function, and $h'$ as an element-wise nonlinear function (\textit{e.g.}, $\tanh$).

\subsection{Model Complexity}

\medskip
\noindent \textbf{Computational Complexity.}
\realite is based on a pipeline that processes the dataset's numeric literals. In particular, the dataset's triples (let $N_t$ be their number) are traversed once (complexity: $O(N_t)$) to fill the matrices $\mathbf{Lh}$ and $\mathbf{Lt}$.
Then, the values of each attribute are aggregated for every relation (complexity: $O(|E|\cdot|A|\cdot|R|)$). So, the overall computational complexity of the preprocessing step is $O(N_t + |E|\cdot|A|\cdot|R|)$.
During training, the computational overhead introduced by the function $g$ is one matrix multiplication and one vector addition in the case of $g_{lin}$, and three matrix multiplications and one vector addition in the case of $g_{gated}$. There is an additional overhead of one matrix multiplication and one vector addition if the learnable function $\mathbf{y}$ is used to aggregate the literals.

\medskip
\noindent \textbf{Trainable Parameters.}
\realite adds some trainable parameters and thus more complexity to the vanilla KGE method (Table~\ref{tab:parameters}). The exact increase of parameters is determined by the function $g$ and the literal aggregation function $\mathbf{\Phi}$. As for the former choice, the additional number of parameters corresponds to the dimensions of the weight matrices $\mathbf{W}$ involved in either Eq.~\ref{eq:glin} or Eq.~\ref{eq:ggated}. Regarding the aggregation method selection, the only case of increased complexity is using the learnable function $\mathbf{y}$. However, since the number of aggregation types is fixed at 11 
the overhead is a fixed number of parameters determined by the dimensions of $\mathbf{W}_a$ and $\mathbf{b}_1$. The added parameters are compared to LiteralE, the latest literal-aware KGE method that alters the learned embeddings of vanilla KGE methods. The additional complexity over LiteralE is constant, thus negligible.

\medskip
\noindent \textbf{Space Complexity.}
During literal preprocessing, for each relation in the dataset, two matrices ($\mathbf{Lh},\mathbf{Lt}$) are created which leads to storing $2\cdot|E|\cdot|A|\cdot|R|$ numeric values (complexity: $O(|E|\cdot|A|\cdot|R|)$). These matrices are reduced to vectors $\mathbf{l}_h,\mathbf{l}_t$ by aggregating the values across one dimension, during the literal preprocessing phase. This means that after preprocessing and throughout model training and testing, the number of stored values is $2\cdot|A|\cdot|R|$.


\begin{table}[t]
\caption{Number of trainable parameters for LiteralE and for variations of \realite. $B$ represents the number of trainable parameters of base models (\textit{e.g.}, TransE). The parentheses in $(+12)$ indicate that this number is added only if the learnable function $\mathbf{y}$ is used to aggregate literals.}
  \centering
    \setlength{\tabcolsep}{4pt}
  \begin{tabular}{ll}
    \toprule
    \textbf{Model} & \textbf{Parameters} \\
    \midrule
    LiteralE & $B + 2D_e^2 + 2|A|D_e + D_e$ \\
    \midrule
    \realite-$g_{lin}$ & $B + D_r^2 + 2|A|D_r + D_r (+ 12)$ \\
    \midrule
    \realite-$g_{gated}$ &  $B + 2D_r^2 + 4|A|D_r + D_r (+ 12)$ \\
    \bottomrule
  \end{tabular}
  
  \label{tab:parameters}
\end{table}

\section{Experimental Setup}

This section outlines the experimental setup used to evaluate the proposed \realite model. We first describe the datasets used for link prediction and node classification, followed by the models employed for each task. Finally, we present the training and evaluation procedures for both tasks.

\subsection{Datasets}
\label{sec:datasets}

\medskip
\noindent \textbf{Link Prediction Datasets.}
\textbf{FB15k-237}~\cite{toutanovaObservedLatentFeatures2015} is a subset of the Freebase KG~\cite{bollackerFreebaseCollaborativelyCreated2008}, focusing on diverse domains such as movies, actors, awards, sports, and sports teams. Originally derived from the FB15k benchmark, \text{FB15k-237} underwent modifications by eliminating inverse relations to enhance the challenge of link prediction. This adjustment aimed to eliminate easily obtainable triples by reversing training triples. The dataset initially lacked numeric literals. It was  enriched by~\cite{kristiadiIncorporatingLiteralsKnowledge2019a} using a SPARQL endpoint for Freebase.
The data split used in this paper aligns with the latest literal-aware KGE~\cite{kristiadiIncorporatingLiteralsKnowledge2019a,wangAugmentingKnowledgeGraphs2022c}.
\textbf{YAGO15k}~\cite{garcia-duranLearningSequenceEncoders2018a}, stands as a subset derived from 
YAGO~\cite{mahdisoltaniYAGO3KnowledgeBase2013}, which functions as a general-domain knowledge graph. Initially extended with numeric literals by~\cite{liuMMKGMultimodalKnowledge2019}, YAGO15k holds a notable advantage in terms of valid numeric triples compared to FB15k-237. We split the dataset in the same way as~\cite{wangAugmentingKnowledgeGraphs2022c}.
The datasets' metadata is shown in Table~\ref{tab:lp-datasets}.


\medskip
\noindent \textbf{Node Classification Datasets.}
For this downstream task, to provide a direct comparison with recent relevant literature~\cite{preisnerUniversalPreprocessingOperators2023a} we use some multimodal datasets of the kgbench collection~\cite{bloemKgbenchCollectionKnowledge2020}. Specifically, we use dmgfull (monuments in the Netherlands), dmg777k (a subset of dmgfull), and mdgenre (movie data extracted from Wikidata). Their statistics are shown in Table~\ref{tab:classification-datasets}. Although these datasets originally contained data of multiple modalities (e.g., images), we only kept the numbers and dates (converted to the decimal format $\text{YYYY}.\text{MM}\text{DD}$).

\begin{table}[t]
    \caption{Datasets used for link prediction}
    \centering
    \setlength{\tabcolsep}{4pt}

        \begin{tabular}{lrrrrr}
            \toprule
            & \textbf{Entities} & \textbf{Relations} & \textbf{Triples} & \textbf{Attributes} & \textbf{Literals} \\
            \midrule
            \textbf{FB15k-237} & 14,541 & 237 & 310,116 & 116 & 29,220 \\
            \textbf{YAGO15k} & 15,136 & 32 & 138,056 & 7 & 23,520 \\
            \bottomrule
        \end{tabular}
    
    \label{tab:lp-datasets}
\end{table}

\begin{table}[t]
\caption{Datasets used for node classification}
  \label{tab:classification-datasets}
  \centering
    \setlength{\tabcolsep}{4pt}
  \begin{tabular}{lrrrr}
    \toprule
     & \textbf{dmgfull} & \textbf{dmg777k} & \textbf{mdgenre} \\
    \midrule
    \textbf{Entities} & 593,291 & 288,379 & 1,001,791 \\
    \textbf{Relations}  & 62 & 60 & 154 \\
    \textbf{Triples} & 1,850,451 & 777,124 & 1,252,247 \\
    \textbf{Numbers}  & 88,168 & 10,706 & 14,352 \\
    \textbf{Dates}  & -- & -- & 113,463 \\
    \midrule
    \textbf{Classes}  & 14 & 5 & 12 \\
    \textbf{Nodes} & 842,550 & 341,270 & 349,344 \\
    \bottomrule
  \end{tabular}
  
\end{table}

\subsection{Models}
\noindent \textbf{Link Prediction Models.}
Similar to literal-aware KGE model~\cite{wangAugmentingKnowledgeGraphs2022c}, we combine \realite with five models: TransE~\cite{NIPS2013_1cecc7a7}, DistMult~\cite{yangEmbeddingEntitiesRelations2014}, ComplEx~\cite{trouillonComplexEmbeddingsSimple2016}, RotatE~\cite{sunRotatEKnowledgeGraph2019}, and TuckER~\cite{balazevicTuckERTensorFactorization2019b}. Note that ConvE~\cite{dettmersConvolutional2DKnowledge2018a} requires the creation of inverse triples. Since we chose a training strategy that does not require the creation of inverse triples, we
do not provide results for ConvE. 
We conduct experiments to determine the best configuration (including the literal aggregation strategy) for each dataset and KGE model. We report the best results in terms of MRR. We compare our \realite to basic KGE models as well as the state-of-the-art models that incorporate numerical values, i.e. KBLN~\cite{garcia-duranKBLRNEndtoEndLearning2018d}, LiteralE~\cite{kristiadiIncorporatingLiteralsKnowledge2019a}, and KGA~\cite{wangAugmentingKnowledgeGraphs2022c}.



\medskip
\noindent \textbf{Node Classification Models.}
We combine \realite with TransE and DistMult to compare our results with those reported in~\cite{preisnerUniversalPreprocessingOperators2023a}.
Additionally, we utilize the literal preprocessing methods presented in the referenced paper to train and evaluate TransE and DistMult. Specifically, we use two preprocessing strategies: KL-REL+LOF, which removes outliers and bins values based on relation similarity, and DATBIN, which converts dates to timestamps for binning~\cite{preisnerUniversalPreprocessingOperators2023a}.
For simplicity, we refer to this combination of preprocessing strategies as MKGA, which is the acronym of that preprocessing framework. As for the classifiers, we use a Support Vector Machine (SVM)~\cite{boserTrainingAlgorithmOptimal1992} and k-Nearest Neighbors (KNN)~\cite{fixDiscriminatoryAnalysisNonparametric1951}.

\subsection{Training}
\label{sec:training}

\noindent \textbf{Link Prediction Training.}
Since \realite focuses on enriching the relations' embeddings, we use a training method that preserves the relation types in the training dataset without adding synthetic inverse relations, unlike Local Closed World Assumption (LCWA)~\cite{aliBringingLightDark2022}. At the same time, due to the small size of the datasets, it is reasonable to perturb the head or tail part of the training triple using all the entities of the dataset.
Another requirement is explicitly training the model to predict a triple's head and tail. So, we choose to train \realite by scoring heads and tails at once for every true triple, based on the concept initially proposed in \cite{lacroixCanonicalTensorDecomposition2018} as `instantaneous multi-class log-loss' (implemented in PyKEEN as `symmetric LCWA'). The chosen loss function is Cross Entropy loss, which generally yields better results than other loss functions~\cite{ruffinelliYouCANTeach2019}.\looseness=-1

For a training triple $(i, j, k)$, a normalized ground truth tensor $y$, and a score tensor $s$, the loss $\ell_{ijk}$ is given by eq.~\ref{eq:full-loss}. The loss consists of two components: the loss for predictions after perturbing the tail entity $k$ (eq.~\ref{eq:left-loss}) and the loss for predictions after perturbing the head entity $i$ (eq.~\ref{eq:right-loss}).
\begin{align}
\ell_{ijk} = & \ell^{(1)}_{ijk} + \ell^{(3)}_{ijk} \label{eq:full-loss} \\ 
\ell^{(1)}_{ijk} = & -\sum_{k'} y_{ijk'} \log\left(\frac{\exp(s_{ijk'})}{\sum_{k'} \exp(s_{ijk'})}\right) \label{eq:left-loss} \\
\ell^{(3)}_{ijk} = & -\sum_{i'} y_{i'jk} \log\left(\frac{\exp(s_{i'jk})}{\sum_{i'} \exp(s_{i'jk})}\right) \label{eq:right-loss}
\end{align}

\medskip
\noindent \textbf{Node Classification Training.}
The KGE model trained on link prediction provides embeddings for the KG's entities. We treat these embeddings as features for the entities used to train the classification models. To train the SVM classifier, we use the hinge loss, which is designed to maximize the margin between different classes. The kNN classifier is a non-parametric algorithm that does not involve explicit training with a loss function.

\subsection{Evaluation Settings}
\label{sec:eval-settings}

\medskip
\noindent \textbf{Overall Link Prediction.} 
For each test triple, the trained model calculates a score after the head entity is replaced with all the dataset entities. This process is repeated for the tail entity. Then, traditional rank-based metrics (i.e., MRR, Hits@1, Hits@10) are calculated based on the position of the test triple in the sorted list of scored triples.

\medskip
\noindent \textbf{Relation-focused Link Prediction.} 
Here the test set is split into groups based on their relation type(s), and then the MRR is calculated.
The splitting is done in two ways: (1) based on the frequency of the relations in the training set (Table~\ref{tab:macro-results-frequency}), to see the performance on long-tail relations and (2) based on the correlation of numerical attributes between the head and tail entities for each relation in the training set (Table~\ref{tab:macro-results-corr}), to see if such correlation makes a difference. For (1), we use 2817 (2.55\% of training triples) as a frequency threshold for the division of relations.
For (2), we calculate the pairwise Pearson's correlation coefficient (denoted by `$corr.\ coef.$') between head and tail attributes and set a threshold of 0.2 for the division of relations.
The rationale behind these groupings is explained in the Appendix.

\medskip
\noindent \textbf{Node Classification.}
All the test entities are labeled using the trained classification model, and then the numbers of true positives, false positives, and false negatives are computed. Afterward, these numbers are used to calculate the micro-F1 score (i.e., the harmonic mean of precision and recall), which is commonly used as a performance measure in classification tasks. For the micro-F1 score, all test samples are treated equally regardless of their class.

\section{Results}

This section presents the experimental results of combining \realite with various vanilla KGE models. We first evaluate the performance of \realite on the link prediction task, comparing it against baseline models and state-of-the-art approaches. Subsequently, we investigate the effectiveness of \realite on the node classification task, analyzing its impact on downstream applications.

\subsection{Link Prediction Results}
\label{sec:lp-results}

\noindent \textbf{Overall Link Prediction.}
Under the standard setting, we observe (Table~\ref{tab:lp-results}) that \realite outperforms the baselines on most backbone KGEs.
For FB15k-237, the maximum percentage increase in MRR when using \realite over using the vanilla KGE models is 17\% (for ComplEx), and the minimum is 6\% (for RotatE).
For YAGO15k, the respective increases are 10\% (for ComplEx) and 7\% (for the rest of the base models).
Furthermore, the best results for YAGO15k regarding MRR and Hits@1 are obtained with \transerealite.
Since YAGO15k has higher quality numeric literals than FB15k237, we consider the results on YAGO15k to be more indicative.
Enhanced by \realite, the performance of almost all vanilla KGE models is improved (except for TuckER on FB15k-237). 
In contrast, LiteralE yields lower metric values than vanilla TransE and TuckER for both datasets; and ComplEx for one dataset.
The performance gain by \realite is the largest for ComplEx, in which each part (\textit{i.e.}, imaginary and real) of the relation embedding is separately enhanced with numeric literals. This characteristic of ComplEx means that embedding-enriching such as \realite and LiteralE are applied twice for this model. Combined with the above observation, this shows the advantage of using the relation-centric \realite versus other methods such as the entity-centric LiteralE.



\begin{table}[t]
  \caption{Link prediction results. In this context, \realite uses $g_{lin}$ as it yielded better results than $g_{gated}$. 
  The baseline results are sourced from the current state-of-the-art paper, with the best results per metric and base KGE model in bold.
  }
  \label{tab:lp-results}
  \centering
    \setlength{\tabcolsep}{4pt}

  \begin{tabular}{lc@{\hspace{4pt}}c@{\hspace{4pt}}cc@{\hspace{4pt}}c@{\hspace{4pt}}c}
    \toprule
    \multicolumn{1}{c}{} &
    \multicolumn{3}{c}{\textbf{FB15k-237}} &
    \multicolumn{3}{c}{\textbf{YAGO15k}} \\
    \midrule
    \multicolumn{1}{c}{\textbf{Model}} & \textbf{MRR} & \textbf{H@1} & \textbf{H@10} & \textbf{MRR} & \textbf{H@1} & \textbf{H@10} \\
    \midrule
    \textbf{TuckER} & 0.354 & 0.263 & 0.536 & 0.433 & 0.360 & 0.571 \\
    +LiteralE & 0.353 & 0.262 & 0.536 & 0.421 & 0.348 & 0.564 \\
    +KBLN & 0.345 & 0.253 & 0.530 & 0.420 & 0.349 & 0.556 \\
    +KGA & \textbf{0.357} & \textbf{0.265} & \textbf{0.540} & 0.454 & 0.380 & 0.592 \\
    +\realite & 0.347 & 0.254 & 0.533 & \textbf{0.463} & \textbf{0.387} & \textbf{0.608} \\ 
    \midrule
    \textbf{TransE} & 0.315 & 0.217 & 0.508 & 0.459 & 0.376 & 0.615 \\ 
    +LiteralE & 0.315 & 0.218 & 0.504 & 0.458 & 0.376 & 0.612 \\ 
    +KBLN & 0.308 & 0.210 & 0.496 & 0.466 & 0.382 & 0.621 \\
    +KGA & 0.321 & 0.223 & 0.516 & 0.470 & 0.387 & \textbf{0.623} \\
    +\realite & \textbf{0.335} & \textbf{0.242} & \textbf{0.520} & \textbf{0.491} & \textbf{0.422} & 0.620 \\ 

    \midrule
    \textbf{DistMult} & 0.295 & 0.212 & 0.463 & 0.457 & 0.389 & 0.585 \\
    +LiteralE & 0.309 & 0.223 & 0.481 & 0.462 & 0.396 & 0.587 \\
    +KBLN & 0.302 & 0.220 & 0.470 & 0.449 & 0.377 & 0.581 \\
    +KGA & 0.322 & 0.233 & 0.502 & 0.472 & 0.402 & 0.606 \\
    +\realite & \textbf{0.340} & \textbf{0.248} & \textbf{0.525} & \textbf{0.489} & \textbf{0.419} & \textbf{0.622} \\ 
    \midrule
    \textbf{ComplEx} & 0.288 & 0.205 & 0.455 & 0.441 & 0.370 & 0.572 \\
    +LiteralE & 0.295 & 0.212 & 0.462 & 0.443 & 0.375 & 0.570 \\
    +KBLN & 0.293 & 0.213 & 0.451 & 0.451 & 0.380 & 0.583 \\
    +KGA & 0.305 & 0.219 & 0.478 & 0.453 & 0.380 & 0.591 \\
    +\realite & \textbf{0.338} & \textbf{0.244} & \textbf{0.528} & \textbf{0.487} & \textbf{0.417} & \textbf{0.621}  \\  
    \midrule
    \textbf{RotatE} & 0.324 & 0.232 & 0.506 & 0.451 & 0.370 & 0.605 \\
    +LiteralE & 0.329 & 0.237 & 0.512 & 0.475 & 0.400 & 0.619 \\
    +KBLN & 0.314 & 0.222 & 0.500 & 0.469 & 0.393 & 0.613 \\
    +KGA & 0.335 & 0.242 & 0.521 & 0.473 & 0.392 & \textbf{0.626} \\
    +\realite & \textbf{0.344} & \textbf{0.249} & \textbf{0.535} & \textbf{0.483} & \textbf{0.409} & 0.624 \\ 
    \bottomrule
  \end{tabular}
\end{table}

\medskip
\noindent \textbf{Relation-focused Link Prediction.}
To examine the performance of literal-aware models for the various relation types, we perform link prediction evaluation after filtering the test set of YAGO15k depending on specific relation groups.

In Table~\ref{tab:macro-results-frequency}, we divide the test triples based on the frequency of their relations in the training set. The frequency is the number of training triples that contain the relation.
We observe that \realite brings performance increase compared to the baselines, especially significant in group ``Long-tail'' (except TuckER). This indicates that \realite is better at modeling the long-tail relations, which constitute the majority in KGs.
We estimate the exception of TuckER+KGA is because the synthetic relations and entities created by KGA lead to a KG structure that benefits TuckER more than other KGE models, as discussed in~\cite{mblum-etal-2024-literalevaluation}. For details, see Appendix.




In Table~\ref{tab:macro-results-corr}, we divide the test triples into groups based on the correlation between the head and tail attributes of their relation in the training set.
We observe that \realite brings performance gain for triples with less correlated literals (``Less corr. lit.''), this indicates \realite is better for capturing nuanced correlations.
In addition, for the triples with more correlated literals (``Corr. lit.''), \realite can significantly boost the performance for some methods (especially TransE enjoys an increase of 47\% and ComplEx 28\%). This means \realite with the mechanism of infusing literals in relations can be very beneficial for some KGE methods.
While for TuckER and RotatE, we can see that KGA leads to higher MRR than \realite in group ``Corr. lit.''. 
We attribute these models' better performance to the additional relations and entities provided by KGA, which might not suit more complex scenarios as discussed in~\cite{mblum-etal-2024-literalevaluation}. For details, see Appendix.

\begin{table}[t]
        \caption{MRR for the filtered test set of YAGO15k based on relation types grouped by their frequency. ``Long-tail'' refers to the test triples that contain relations present in $\leq 2.55\%$ of the training triples. ``Frequent'' refers to the remaining test triples. ``All triples'' refers to all test triples. The best results per base model per triple group are in bold. 
        The percentages indicate the MRR increase for the top literal-extended version of a base model versus the second-best per triple group.
        }
        \label{tab:macro-results-frequency}
    \centering
    \setlength{\tabcolsep}{4pt}

            \begin{tabular}{l@{\hspace{4pt}}lrrr}
                \toprule
                \textbf{Model} &  & \textbf{Frequent} & \textbf{Long-tail} & \textbf{All triples} \\
                \midrule
                \textbf{TuckER} 
                &+LiteralE & 0.464 & 0.282 & 0.441 \\
                &+KGA & 0.480 & +5\% \textbf{0.288} & 0.455 \\
                &+\realite & +2\% \textbf{0.491} & 0.274 & +2\% \textbf{0.463} \\
                \midrule
                \textbf{TransE} 
                &+LiteralE & 0.489 & 0.233 & 0.456 \\
                &+KGA & 0.502 & 0.237 & 0.469 \\
                &+\realite & +5\% \textbf{0.525} & +11\% \textbf{0.262} & +5\% \textbf{0.491} \\
                \midrule
                \textbf{DistMult} 
                &+LiteralE & 0.492 & 0.250 & 0.461 \\
                &+KGA & 0.501 & 0.258 & 0.470 \\
                &+\realite & +4\% \textbf{0.519} & +10\% \textbf{0.284} & +4\% \textbf{0.489} \\
                \midrule
                \textbf{ComplEx} 
                &+LiteralE & 0.470 & 0.253 & 0.442 \\
                &+KGA & 0.479 & 0.256 & 0.451 \\
                &+\realite & +7\% \textbf{0.514} & +17\% \textbf{0.300} & +8\% \textbf{0.487} \\
                \midrule
                \textbf{RotatE} 
                &+KGA & 0.494 & 0.298 & 0.469 \\
                &+\realite & +3\% \textbf{0.510} & +1\% \textbf{0.302} & +3\% \textbf{0.483} \\
                \bottomrule
            \end{tabular}


    \vspace{-2ex}
    \end{table}
    
    \begin{table}[t]
            \caption{MRR for the filtered test set of YAGO15k based on relation types grouped by literal correlation. ``Corr. lit.'' refers to the test triples where the head and tail attributes are more correlated (the relation has at least one pair of head and tail attributes with $|corr.\ coef.| \geq 0.2$ in the training set). ``Less corr. lit.'' refers to the remaining test triples where the head and tail attributes are less correlated. ``All triples'' refers to all test triples. The best results per base model per triple group are in bold. 
        The percentages indicate the MRR increase for the top literal-extended version of a base model versus the second-best per triple group.
        }
        \label{tab:macro-results-corr}
        \centering
        \setlength{\tabcolsep}{4pt}

            \begin{tabular}{l@{\hspace{4pt}}lrrr}
                \toprule
                \textbf{Model} & &\textbf{Less corr. lit.} & \textbf{Corr. lit.} & \textbf{All triples} \\
                \midrule
                \textbf{TuckER} & 
                +LiteralE & 0.465 & 0.272 & 0.441 \\
                &+KGA & 0.477 & +10\% \textbf{0.298} & 0.455 \\
                &+\realite & +2\% \textbf{0.488} & 0.289 & +2\% \textbf{0.463} \\
                \midrule
                \textbf{TransE} 
                &+LiteralE & 0.500 & 0.146 & 0.456 \\
                &+KGA & 0.510 & 0.176 & 0.469 \\
                &+\realite & +3\% \textbf{0.524} & +47\% \textbf{0.259} & +5\% \textbf{0.491} \\
                \midrule
                \textbf{DistMult} 
                &+LiteralE & 0.494 & 0.227 & 0.461 \\
                &+KGA & 0.501 & 0.245 & 0.470 \\
                &+\realite & +3\% \textbf{0.517} & +17\% \textbf{0.287} & +4\% \textbf{0.489} \\
                \midrule
                \textbf{ComplEx} 
                &+LiteralE & 0.477 & 0.193 & 0.442 \\
                &+KGA & 0.479 & 0.245 & 0.451 \\
                &+\realite & +7\% \textbf{0.511} & +28\% \textbf{0.314} & +8\% \textbf{0.487} \\
                \midrule
                \textbf{RotatE} 
                &+KGA & 0.488 & +12\% \textbf{0.333} & 0.469 \\
                &+\realite & +4\% \textbf{0.509} & 0.296 & +3\% \textbf{0.483} \\
                \bottomrule
            \end{tabular}

\end{table}

\medskip
\noindent \textbf{Analysis of Literal Aggregation Methods.}
For the link prediction training, during \realite's hyperparameter optimization (HPO), the choices for literal aggregation were $\{mean, median, mode, min, \mathbf{y}\}$. We did not include the rest of the
provided aggregation types
to limit the HPO search space.
Besides, the provided learnable function $\mathbf{y}$ combines all supported aggregation types.
Based on Table~\ref{tab:best-configs-aggr}, the choice of aggregation function that leads to optimal results generally depends on both the dataset and the model. For TransE, RotatE, and TuckER, the aggregation function is the same in both YAGO15k and FB15k-237. In addition, $mode$ is found in 50\% of the configurations that yielded the best results per dataset per model, while $mean$ and $min$ in 20\%. The least used aggregation function was the learnable function $\mathbf{y}$, which is present in 10\% of the configurations. $median$ was the only function among the HPO choices that was not selected as part of any configuration. Further analysis is included in the Appendix.

\begin{table}[hb]
    \caption{Aggregation function used in the best hyperparameter configurations of \realite. Function $\mathbf{y}$ is a learnable combination of all supported aggregation types (Eq.~\ref{aggr-combination}).}
    \label{tab:best-configs-aggr}
    \centering
    \setlength{\tabcolsep}{4pt}
    \begin{tabular}{lccccc}
        \toprule
         & \textbf{TransE} & \textbf{DistMult} & \textbf{ComplEx}  & \textbf{RotatE} & \textbf{TuckER} \\
        \midrule
        \textbf{FB15k-237}& $mode$ & $\mathbf{y}$ & $mode$ & $mode$ & $mean$ \\ 
        \midrule
        \textbf{YAGO15k} &$mode$ & $min$ & $min$ & $mode$ & $mean$ \\ 
        \bottomrule
    \end{tabular}
\end{table}

\subsection{Node Classification Results}

In Table~\ref{tab:classification-results}, we present an evaluation of KGE models on the node classification datasets. It is apparent that for every dataset, the best result is obtained using \realite. In particular, for dmg777k, there is a vast increase in F1 score when using \realite compared to the vanilla KGE models. On average, for all combinations of the KGE model and classification model on dmg777k, this increase is 16\%. We can observe that when using TransE with KNN or SVM on dmgfull and with KNN on mdgenre, \realite seems to perform worse than the vanilla TransE. However, even in those cases, \realite outperforms MKGA, the literal-aware baseline. In addition, across all used datasets and KGE base models, \realite yields better results when paired with the SVM classifier than when paired with KNN.

\begin{table}[t]
\caption{Node classification results -- Micro-F1 score. In this context, \realite uses $g_{gated}$ as it yielded better results than $g_{lin}$.
  The best results per dataset are underlined. The best results per base KGE model per classifier are in bold.
  }
  \centering
    \setlength{\tabcolsep}{4pt}
  \begin{tabular}{lc@{\hspace{4pt}}cc@{\hspace{4pt}}cc@{\hspace{4pt}}cc@{\hspace{4pt}}c}
    \toprule
    \multicolumn{1}{c}{} &
    \multicolumn{2}{c}{\textbf{dmg777k}} &
    \multicolumn{2}{c}{\textbf{dmgfull}} &
    \multicolumn{2}{c}{\textbf{mdgenre}} \\
    \midrule
    \multicolumn{1}{l}{\textbf{Model}}  & \textbf{KNN} & \textbf{SVM} & \textbf{KNN} & \textbf{SVM} & \textbf{KNN} & \textbf{SVM} \\
    \midrule
    \textbf{TransE} & 0.506 & 0.528 & \textbf{0.649} & \textbf{0.662} & \textbf{0.634} & 0.646 \\
    +MKGA  & 0.439 & 0.472 & 0.583 & 0.573 & 0.607 & 0.606 \\
    +\realite  & \textbf{0.609} & \textbf{0.638} & 0.597 & 0.605 & 0.632 & \textbf{0.662} \\    
    \midrule
    \textbf{DistMult} & 0.548 & 0.542 & 0.619 & 0.658 & 0.605 & 0.622 \\
    +MKGA & 0.472 & 0.495 & 0.605 & 0.647 & 0.630 & 0.640 \\
    +\realite & \textbf{0.582} & \textbf{\underline{0.642}} & \textbf{0.639} & \textbf{\underline{0.682}} & \textbf{0.659} & \textbf{\underline{0.674}} \\ 
    \bottomrule
  \end{tabular}
  
  \label{tab:classification-results}
\end{table}

\section{Conclusion}


This paper introduces \realite, a novel approach to enhance Knowledge Graph Embedding (KGE) with literal information. \realite enhances traditional KGEs by incorporating information from the relevant literals of the head and tail entities into the relation embedding. We present variants of \realite with different methods for aggregating numeric literals, including an automated learning approach to combine different aggregations. 
The experiments demonstrate the superiority of \realite over state-of-the-art methods for literal-aware embeddings, achieving comparable or improved performance in downstream tasks. Notably, detailed analyses reveal that \realite excels in scenarios involving numeric literals with higher correlation and long-tail relations, showcasing its versatility and efficacy in capturing nuanced relationships within knowledge graphs.

\subsubsection{\ackname}
The work was partially supported by EU projects Graph Massivizer (GA 101093202), Dome 4.0 (GA 953163), SMARTY (GA 101140087), and enRichMyData (GA 101070284).


%
%
%
\bibliographystyle{splncs04}
\bibliography{references}
%

\end{document}